\pdfoutput=1
\documentclass[11pt]{article}
\usepackage[]{acl}
\usepackage[greek,english]{babel}
\usepackage{xurl}
\usepackage{amsmath}
\usepackage{times}
\usepackage{placeins}
\usepackage{latexsym}
\usepackage{array}
\usepackage[capitalise]{cleveref}
\usepackage[T1]{fontenc}
\usepackage[utf8]{inputenc}

\usepackage{enumitem}
\usepackage{microtype}
\usepackage{xspace}
\usepackage{graphicx}
\usepackage{booktabs}
\usepackage{siunitx}
\usepackage{microtype}
\newcommand{\philta}{\textsc{PhilTa}\xspace}
\newcommand{\philberta}{\textsc{PhilBERTa}\xspace}
\newcommand{\sphilberta}{\textsc{SPhilBERTa}\xspace}
\newcommand{\roberta}{\textsc{RoBERTa}\xspace}
\newcommand{\bert}{\textsc{BERT}\xspace}
\newcommand{\sbert}{\textsc{SBERT}\xspace}
\newcommand{\electra}{\textsc{ELECTRA}\xspace}
\newcommand{\mpnet}{\textsc{MPNet}\xspace}

\usepackage{numprint}
\npthousandsep{\,}
\usepackage{pifont}
\usepackage[dvipsnames]{xcolor}
\newcommand{\cmark}{\textcolor{green!20!black!70}{\ding{51}}}
\newcommand{\xmark}{\textcolor{red}{\ding{55}}}
\usepackage{csquotes}
\usepackage{adjustbox}
\usepackage{calc}
\newcommand{\citeposs}[1]{\citeauthor{#1}'s (\citeyear{#1})}
\usepackage{float}
\title{\textit{Graecia capta ferum victorem cepit}\\Detecting Latin Allusions to Ancient Greek Literature}

\author{Frederick Riemenschneider
    \hspace{2em}
    Anette Frank\\
  Department of Computational Linguistic\\
  Heidelberg University\\
  69120 Heidelberg, Germany \\
  {\tt \{riemenschneider|frank\}@cl.uni-heidelberg.de}
}

\begin{document}
\maketitle
\begin{abstract}
Intertextual allusions hold a pivotal role in Classical Philology, with Latin authors frequently referencing Ancient Greek texts. Until now, the automatic identification of these intertextual references has been constrained to monolingual approaches, seeking parallels solely within Latin or Greek texts. In this study, we introduce \sphilberta, a trilingual Sentence-\roberta model tailored for Classical Philology, which excels at cross-lingual semantic comprehension and identification of identical sentences across Ancient Greek, Latin, and English. We generate new training data by automatically translating English texts into Ancient Greek.  Further, we present a case study, demonstrating \sphilberta's capability to facilitate automated detection of intertextual parallels. Our models and resources are available at \url{https://github.com/Heidelberg-NLP/ancient-language-models}.
\end{abstract}
\section{Introduction}

The study of intertextuality and allusions to literary sources has a longstanding tradition in Classical Philology, highlighting complex interconnections between different literary works. 
During the 1960s, the concept of intertextuality was shaped by a comprehensive theoretical framework developed by scholars such as Julia Kristeva, Ferdinand de Saussure, and Michail Bakhtin. The term \enquote{intertextuality} itself was introduced by Kristeva during this pivotal era \citep{surveyinter,Bendlin2006,kristeva1986kristeva,orr2003intertextuality}.

Intertextuality proves particularly crucial when examining Roman literature's relationship with Ancient Greek texts. Many Latin authors consciously mirrored elements of Greek classics, making intertextuality an essential concept for understanding this cultural literary exchange.\footnote{Cf.~\citet{hutchinson}: \enquote{How Latin literature relates to Greek literature is one of the most fundamental questions for Latin literature, and for the reception of Greek.}}

The importance of intertextuality, especially given the considerable attention it has received, is beyond dispute. While there exists a plethora of theoretical work exploring specific forms of intertextuality, our focus in this work is on the general occurrence of textual resemblances, specifically within Latin and Greek texts. 

Traditionally, the identification of such parallels has largely relied on scholars' close reading. However, recent years have seen the development of statistical NLP tools -- driven especially by the Tesserae project \citep{coffee2012tesserae,tesserae2} 
at the forefront of this movement --
that are able to automatically uncover a considerable number of textual parallels.
These approaches, however, typically rely on string-level parallels and are grounded in carefully designed rules and scoring functions. Notably, these systems
are generally restricted to detecting parallels in the same language, as they rely on identifying identical tokens or stems.

Recently, the breakthrough in self-supervised training of powerful pre-trained language  models (PLMs) has also led to a surge of diverse PLMs for Classical Philology
\citep{bamman2020latin,yamshchikov-etal-2022-bert,mercelis-keersmaekers-2022-electra,singh-etal-2021-pilot,riemenschneider2023exploring}.
In fact, two recent case studies in \citet{bamman2020latin} and \citet{burns2023latincy} have shown that contextualized embeddings produced by such models can indeed identify texts bearing similar content.
While a rigorous quantitative evaluation of these findings still remains to be conducted, the perceived potential of using these models for finding intertextual relations is clearly sparking widespread interest.

However, research into modern language ana\-lysis tasks has demonstrated that sentence embeddings derived solely from standalone \bert- or \roberta-based models 
generate suboptimal and inefficient embeddings. This insight
led to the creation of Sentence-\bert (\sbert) models \citep{reimers-gurevych-2019-sentence}.

Among the latest language models introduced in the field of Classical Philology is \philberta \citep{riemenschneider2023exploring}, a \roberta-based model pre-trained on Ancient Greek, Latin, and English language data. Building upon this model, we present \sphilberta, a model tailored to
the discovery of intertextual parallels across Latin, Ancient Greek, and English texts.

In this work, our goal is to move away from 
systems relying on hand-crafted rules, and instead to employ state-of-the-art tools for identifying intertextual relations that are easy to adapt to a wide variety of languages from Classical Philology and beyond. 
Most importantly, we probe the feasibility of uncovering intertextual parallels \emph{across languages}, an area that has been largely neglected in the automatic identification of intertextual allusions until this point. This novel capability will considerably enlargen the space for new findings, by being able to compare texts directly across languages.

We show that \sphilberta is proficient in recognizing direct translations of sentences in Ancient Greek, Latin, and English, thereby demonstrating comprehensive cross-lingual competence. Applying our model directly to texts of philological significance not only underlines its practical applicability but also highlights areas for improvement, suggesting promising avenues for future exploration.

In summary, our contributions are as follows:

\begin{enumerate}[label=\roman*), noitemsep]
\item We introduce \sphilberta, a multilingual sentence transformer
for Latin, Ancient Greek, and English. To our knowledge, we are the first to apply this type of model 
to automatically detect passages of potential \emph{cross-lingual} allusions in Latin texts.
\item To alleviate the scarcity of parallel sentence pairs for training \sphilberta, we augment the available resources by automatically translating English texts to Ancient Greek using an existing multilingual T5 model pre-trained on Ancient Greek, Latin, and English data.
\item We conduct experiments on retrieving translations or similar sentences from textual passages in foreign-language texts, using cross-lingual \sphilberta sentence embeddings.
\item Our experiments demonstrate that \textsc{SPhil\-BERTa} is able to detect translations with high accuracy and that data augmentation significantly enhances the performance of the system for Ancient Greek. While finding textual allusions still requires philological expertise, we present cases where the model identifies passages linked to known allusive texts.
\end{enumerate}

\section{Related Work}
\paragraph{Detecting Intertextual Allusions.} Initiated in 2008, the Tesserae project \citep{coffee2012tesserae,tesserae2} has been instrumental in advancing the automatic detection of intertextuality in Latin and Greek texts. Their open-source tools have seen numerous enhancements and refinements over the years.\footnote{\url{https://tesserae.caset.buffalo.edu/blog/about-tesserae/}.}

Existing research has explored matching words or stems \citep{coffee2012tesserae} as well as methods that focus on semantics \citep{sense}. Additionally, techniques that combine both lexical and semantic elements have been examined, where semantic understanding is established through word embeddings \citep{manjavacas-etal-2019-feasibility} or via the (Ancient Greek) WordNet \citep{bizzoni-etal-2014-making}. While the majority of preceding studies have concentrated on detecting text reuse in the Bible and various religious texts,  \citet{burns-etal-2021-profiling} focus on Classical Latin literature.

However, to our knowledge, no efforts have been undertaken to automatically detect intertextual similarities across languages, specifically between Greek, Latin, and English texts.
This lack is likely due to the inherent complications of inducing cross-language mappings, a difficulty that arises both with surface form-based strategies and with techniques utilizing word
embeddings. Notwithstanding, this gap is of significant importance, as it overlooks the frequent appearance of such allusions, especially from Latin to Greek literature.

\paragraph{Language Models for Classical Philology.} 
\citet{bamman2020latin} and \citet{mercelis-keersmaekers-2022-electra} introduced Latin \bert and \electra models, respectively. For Ancient Greek, \citet{singh-etal-2021-pilot} and \citet{yamshchikov-etal-2022-bert} provided \bert models, initialized from Modern Greek \bert and subsequently trained on Ancient Greek data.  Similarly, the UGARIT project has successfully explored the usage of the XLM-R model \citep{conneau-etal-2020-unsupervised} for Ancient Greek and Latin texts \citep{yousef-etal-2022-automatic,yousef-etal-2022-automatic-translation}, even though XLM-R has not been pre-trained on Ancient Greek texts.  Recently, \citet{riemenschneider2023exploring} have complemented the encoder-only landscape with encoder-decoder models and developed trilingual models using Ancient Greek, Latin, and English texts. Moreover, \citet{kostkan-etal-2023-odycy} and \citet{burns2023latincy} have developed odyCy and la\-tinCy, respectively, as dedicated spaCy pipelines\footnote{\url{https://spacy.io/}.} for Ancient Greek and Latin.
 
\paragraph{\sbert Embeddings.} 
\citet{reimers-gurevych-2019-sentence} have shown that vanilla \bert embeddings are not suitable for creating sentence embeddings,  and instead proposed the S(entence)-\bert models, which are based on Siamese and triplet network structures. Building on their work, \citet{reimers-gurevych-2020-making} introduced a method to learn multilingual sentence embeddings via multilingual knowledge distillation. This method realizes knowledge transfer from a monolingual teacher model to a student model, 
by training the student model to align the original sentence and its translation to the same location in the embedding space.

\section{Methodology}
\label{sec:methodology}
We closely follow \citeposs{reimers-gurevych-2020-making} multilingual knowledge distillation recipe. Their method requires a monolingual teacher model \(M\) and parallel sentences in the given source language and the target language(s) \(((s_1, t_1), ..., (s_n, t_n))\).

The teacher trains a student model \(\hat{M}\) such that \(\hat{M}(s_i) \approx M(s_i)\) and \(\hat{M}(t_i) \approx M(s_i)\). For a given mini-batch \(\mathcal{B}\), the mean-squared loss is minimized:

\begin{equation*}
\resizebox{\linewidth}{!}{\(\frac{1}{|\mathcal{B}|} \sum_{j \in \mathcal{B}} \left[ (M(s_j) - \hat M(s_j))^2 +  (M(s_j) - \hat M(t_j))^2 \right]\)\,.}
\end{equation*}
In other words, the student model is trained to map a given sentence to the same vector across languages, i.e., the translation of a given sentence should be mapped to the same vector as the source sentence. Notably, this method is not restricted to a bilingual setup. Instead, the student can be trained to map sentence vectors stemming from multiple languages to the same vector, namely the one provided by the teacher model.

In our work, the teacher and student \sbert models to be used for cross-lingual knowledge transfer will be initialized from strong transformer language models for the respective languages.
For the English teacher model, we build on the \mpnet model of \citet{song2020mpnet}, an encoder-only model that has been pre-trained using a combination of masked language modeling and permuted language modeling. Specifically, we use different sentence transformer variants induced from \mpnet, as provided by the SBERT library \citep{reimers-gurevych-2019-sentence}. 
For the student model, we experiment with initializing it from different multilingual models: XLM-R \citep{conneau-etal-2020-unsupervised}, a multilingual model based on \roberta that covers 100 languages, including Modern Greek and Latin, in contrast to \philberta \citep{riemenschneider2023exploring}, a recent trilingual model that has been pre-trained on Ancient Greek, Latin, and English texts.

More detail about our models and the specific experimental setup is provided in \cref{sec:experiments}.

\section{Parallel Data}
\label{sec:data}
As outlined in \cref{sec:methodology}, the knowledge distillation method of \citet{reimers-gurevych-2020-making}
crucially depends on the availability of parallel sentences between the relevant source and target languages -- here, the source language English for the teacher model, and English, Ancient Greek, and Latin for our student model. 

We collect this data from various sources: from the Perseus Digital Library,\footnote{\url{https://github.com/PerseusDL/canonical-greekLit} and \url{https://github.com/PerseusDL/canonical-latinLit}.} from parallel Bible data,\footnote{\url{https://github.com/npedrazzini/parallelbibles/tree/main}.} parallel English-to-Greek sentences from the OPUS corpus \citep{tiedemann-2012-parallel}, and an extensive collection of parallel English and Latin sentences available on the Huggingface Hub.\footnote{\url{https://huggingface.co/datasets/grosenthal/latin_english_parallel}.} We refer to the latter dataset as \enquote{Rosenthal}, named after its associated account. 

The Perseus project features a large collection of Ancient Greek and Latin texts, many of them with corresponding translations. However, the alignment of the provided data is not always fine-grained enough for our purpose. Therefore, we align individual lines with their corresponding translation, and discard lines that we cannot align successfully.

To generate additional parallel data for enhanced knowledge transfer, we experiment with translating the English portions of the Rosenthal dataset, which consists solely of English and Latin parallel data, into Ancient Greek. This required first fine-tuning the multilingual \philta model\footnotemark{} on the Perseus data to enable translation from English to Ancient Greek. Subsequently, we used the trained \textsc{PhilTa}$_{\text{En}\rightarrow\text{Grc}}$ model to translate the Rosenthal dataset into Ancient Greek, thereby expanding it to a trilingual parallel dataset.

\begin{table}
\begin{tabular}{lrrr}
		& \textbf{English} & \textbf{Greek} & \textbf{Latin} \\
		\midrule
		\textbf{Perseus} & \numprint{3743}K & \numprint{2120}K & \numprint{384}K \\
		\textbf{Bible} & \numprint{897}K & \numprint{128}K &  \numprint{520}K\\
		\textbf{Opus} & \numprint{5}K & \numprint{4}K & ---\\
		\textbf{Rosenthal} & \numprint{3428}K & \numprint{2370}K\(^\dagger\) & \numprint{2095}K \\
\end{tabular}
\caption{Dataset statistics (in number of words) of available parallel sentences across languages. The Greek Rosenthal data marked with a dagger (\(\dagger\)) has been translated using \textsc{PhilTa}$_{\text{En} \rightarrow \text{Grc}}$.\footnotemark[\value{footnote}]}
\label{tab:stats}
\end{table}

\newlength{\enla}
\settowidth{\enla}{\widthof{En\(\rightarrow\)La}}
\newlength{\engrc}
\settowidth{\engrc}{\widthof{En\(\rightarrow\)Grc$^\dagger$}}
\begin{table*}
    \centering
    \resizebox{\linewidth}{!}{
    \begin{tabular}{>{\raggedright}p{4cm}>{\raggedright}p{2cm}>{\centering}p{2cm}>{\raggedleft}p{1.25cm}>{\raggedleft}p{1.25cm}>{\raggedleft}p{1.25cm}>{\raggedleft}p{1.25cm}>{\raggedleft}p{1.25cm}r}
        \textbf{Teacher} & \textbf{Student} & \textbf{\philta-} &\multicolumn{2}{c}{\textbf{Bible}} & \multicolumn{2}{c}{\textbf{Perseus}} & \multicolumn{2}{c}{\textbf{Rosenthal}} \\
        &&\textbf{translations}&En\(\rightarrow\)La&La\(\rightarrow\)En&En\(\rightarrow\)La&La\(\rightarrow\)En&En\(\rightarrow\)La\hphantom{-\enla + \engrc}&La\(\rightarrow\)En\hphantom{-\enla + \engrc}\\
        \midrule
        \texttt{all-mpnet-base-v2} & XLM-R  & \xmark & \(0.10\) & \(0.10\) & \(0.30\) & \(0.60\) & \(0.50\) & \(0.60\) \\
        \texttt{all-mpnet-base-v2} & \philberta & \xmark &\(96.10\) & \(95.60\) & \(90.10\)& \(88.40\)& \(95.90\) & \(95.20\) \\
        \texttt{multi-qa-mpnet} & \philberta & \xmark & \(\textbf{96.90}\) & \(\textbf{96.00}\) & \(91.60\) & \(\textbf{91.30}\) & \(\textbf{97.90}\) & \(\textbf{96.90}\)\\
        \texttt{multi-qa-mpnet} & \philberta & \cmark  & \(96.40\) & \(95.90\) & \(\textbf{91.90}\) & \(90.90\) & \(97.80\) & \(96.60\) \\
        
    \end{tabular}}
    \caption{Translation accuracy for various \emph{English-Latin} test sets. Utilizing XLM-R as a student model leads to catastrophic results.  It is crucial to substitute \philberta as the student model for successful model training. Switching to the semantically-oriented \texttt{multi-qa-mpnet} from the broader \texttt{all-mpnet-base-v2} provides further enhancements.}
    \label{tab:resultsla}
    \vspace*{7mm}
    
    \centering
    \resizebox{\linewidth}{!}{
    \begin{tabular}{>{\raggedright}p{4cm}>{\raggedright}p{2cm}>{\centering}p{2cm}>{\raggedleft}p{1.25cm}>{\raggedleft}p{1.25cm}>{\raggedleft}p{1.25cm}>{\raggedleft}p{1.25cm}>{\raggedleft}p{1.25cm}r}
        \textbf{Teacher} & \textbf{Student} & \textbf{\philta-} &\multicolumn{2}{c}{\textbf{Bible}} & \multicolumn{2}{c}{\textbf{Perseus}} & \multicolumn{2}{c}{\textbf{Rosenthal}} \\
        &&\textbf{translations}&En\(\rightarrow\)Grc&Grc\(\rightarrow\)En&En\(\rightarrow\)Grc&Grc\(\rightarrow\)En&En\(\rightarrow\)Grc\(^\dagger\)&Grc$^\dagger$\(\noindent\rightarrow\)En\\
        \midrule
        \texttt{all-mpnet-base-v2} & XLM-R  & \xmark & \(0.20\) & \( 0.20\) & \(0.30\) & \(0.10\) & \(0.30\) & \(0.10\)\\
        \texttt{all-mpnet-base-v2} & \philberta & \xmark & \(96.50\) & \(96.50\) & \(89.50\) & \(87.40\) &  \(93.39\) & \(92.49\)\\
        \texttt{multi-qa-mpnet} & \philberta & \xmark & \(97.80\) & \(97.70\) & \(89.80\) & \(88.80\) & \(92.29\)& \(86.99\)\\
        \texttt{multi-qa-mpnet} & \philberta & \cmark  & \(\textbf{98.30}\) & \(\textbf{98.00}\) & \(\textbf{91.10}\) & \(\textbf{90.50}\) & \(\textbf{96.80}\) & \(\textbf{94.29}\)\\
        
    \end{tabular}}
    \caption{Translation accuracy for various \emph{English-Greek} test sets. The Greek Rosenthal data has been translated by \philta. We see the same trends as in \cref{tab:resultsla}. The enrichment of the training corpus with additional \philta-translated content notably increases the performance for Ancient Greek.}
    \label{tab:resultsgrc}
        \vspace*{7mm}

    \centering
    \resizebox{\linewidth}{!}{
    \begin{tabular}{>{\raggedright}p{4cm}>{\raggedright}p{2cm}>{\centering}p{3.5cm}>{\raggedleft}p{1.5cm}>{\raggedleft}p{1.5cm}>{\raggedleft}p{1.5cm}r}
        \textbf{Teacher} & \textbf{Student} & \textbf{\philta-translations} &\multicolumn{2}{c}{\textbf{Bible}} & \multicolumn{2}{c}{\textbf{Rosenthal}} \\
        &&&La\(\rightarrow\)Grc&Grc\(\rightarrow\)La&La\(\rightarrow\)Grc\(^\dagger\)&Grc\(^\dagger\)\(\rightarrow\)La\\
        \midrule
        \texttt{all-mpnet-base-v2} & XLM-R  & \xmark & \(0.10\) & \(0.10\) & \(0.20\) & \(0.20\)\\
        \texttt{all-mpnet-base-v2} & \philberta & \xmark & \(96.10\) & \(95.60\) & \(83.97\) & \(83.67\)\\
        \texttt{multi-qa-mpnet} & \philberta & \xmark & \(96.50\) & \(96.69\)& \(84.97\)& \(82.57\)\\
        \texttt{multi-qa-mpnet} & \philberta & \cmark  & \(\textbf{96.70}\) & \(\textbf{96.90}\) & \(\textbf{92.08}\) & \(\textbf{91.68}\)\\
        
    \end{tabular}}
    \caption{Translation accuracy for various \emph{Latin-Greek} test sets. The Greek Rosenthal data has been translated by \philta. We see similar trends as described in \cref{tab:resultsla,tab:resultsgrc}.}
    \label{tab:resultsgrcla}
\end{table*}

\footnotetext{\philta \citep{riemenschneider2023exploring} is a trilingual encoder-decoder model based on T5 \cite{JMLR:v21:20-074} that was pre-trained on Ancient Greek, Latin, and English data.}
\cref{tab:stats} provides the data statistics.
Since parts of the corpora overlap, we deduplicate the data.

\section{Experiments}
\label{sec:experiments}

Our first aim is to compare different model configurations. We test the following configurations:

\begin{itemize}
    \item \textbf{Teacher Model.} We use the \texttt{all\--mpnet\--base-v2}\footnote{\url{https://huggingface.co/sentence-transformers/all-mpnet-base-v2}.}  and the \texttt{multi\--qa-mpnet-base-dot-v1}\footnote{\url{https://huggingface.co/sentence-transformers/multi-qa-mpnet-base-dot-v1}.} model from the \sbert library \citep{reimers-gurevych-2019-sentence} as teacher models. While the former is fine-tuned on a variety of tasks, the latter is optimized for semantic search.
    \item \textbf{Student Model.} We compare the performance of XLM-R \citep{conneau-etal-2020-unsupervised} to that of \philberta \citep{riemenschneider2023exploring} when used as student models. XLM-R serves as a well-established multilingual baseline.
    \item \textbf{Data Augmentation.} We evaluate whether the automatic English-to-Greek translations produced by \textsc{PhilTa}$_{\text{En} \rightarrow \text{Grc}}$ to extend the Rosenthal dataset improve task performance.
    
\end{itemize}

In order to transparently evaluate our models, we first measure their ability to correctly detect translations of a sentence.
For each parallel dataset, we hold out \numprint{1000} sentences as test sets.
Given a query, i.e., the embedding of a specific sentence in the source language, we compute the cosine similarity to the embeddings of all \numprint{1000} sentences in the target language.

Following \citet{reimers-gurevych-2020-making}, we measure the success of our models by determining \textit{translation accuracy}: we count a translation to be correctly identified if the model computes the highest cosine similarity between the query and its correct translation, and vice versa.
This evaluates the student model's ability to align a source language sentence with an equivalent target language sentence.

However, our primary interest is whether the model can effectively link Ancient Greek and Latin texts. Regrettably, the volume of parallel data available in Ancient Greek and Latin is severely constrained. Consequently, we utilize Bible data, which is accessible in Ancient Greek, Latin, and English.
Again, we examine the model's performance on \numprint{1000} test sentences, given in Ancient Greek or Latin. We ensure that the model has not encountered any of these sentences in its training data, either in English or Latin, or in Ancient Greek. In addition, we use the \textsc{PhilTa}$_{\text{En} \rightarrow \text{Grc}}$-generated Ancient Greek test set translations of the Rosenthal corpus and compare them to their Latin originals.

We are aware that the task of identifying intertextual allusions poses a much greater challenge than merely recognizing translations, as allusions typically exhibit more subtlety and may extend beyond sentence or verse boundaries. However, we consider this evaluation a transparent method for comparing the effectiveness of different model configurations and an approximate measure to evaluate the potential success of our models in identifying intertextual allusions across languages.

\paragraph{Experiment Details.} We train all models with the exact same configurations. We fine-tune all models for \numprint{30} epochs, using a batch size of 32, the AdamW optimizer with a learning rate of \num{2e-5}, and \numprint{10000} warmup-steps. The best-performing model is selected based on the translation accuracy derived from a total of \numprint{2000} held-out validation examples, comprised of \numprint{1000} English-Greek and \numprint{1000} English-Latin sentence pairs.

\section{Results}
We present our results for the different configurations in \cref{tab:resultsla,tab:resultsgrc,tab:resultsgrcla}. Specifically, 
we evaluate: i) the performance of different \emph{teacher models} (the more general \texttt{all-mpnet-base-v2} \sbert model in comparison to the \texttt{multi-qa-mpnet} \sbert fine-tuned for semantic search), ii) different \emph{student models} (XLM-R versus the \philberta model), and iii) \emph{augmenting the parallel data} for training \sphilberta using \textsc{PhilTa}\(_{{\text{En} \rightarrow \text{Grc}}}\)-translated texts.

Employing XLM-R as the student model leads to catastrophic performance. Specifically, the model never surpasses the 1\% mark in test set performance.  We observed this trend consistently, regardless of the model configuration or the random seed employed. This outcome is, to some degree, to be expected, as XLM-R is not pre-trained on Ancient Greek data. Still, it is surprising that XLM-R performs so badly also on Latin data, as its pre-training corpus contained a Latin portion. Moreover, the UGARIT project \citep{yousef-etal-2022-automatic,yousef-etal-2022-automatic-translation} has successfully adapted XLM-R to Ancient Greek. We hypothesize that the effectiveness of a broadly multilingual but unspecialized model may be task-dependent, and continuing self-supervised pre-training on Ancient Greek texts may be required for XLM-R to adapt adequately. These findings highlight the importance of initializing the student model with a model that is proficient in the target languages.

Initializing the student model with \philberta yields strong performance, often surpassing 95\% translation accuracy. Generally, employing \texttt{multi-qa-mpnet} as a teacher model contributes to a slight performance improvement over \texttt{all-mpnet-base-v2}. Yet, when testing the model on the Ancient Greek Rosenthal corpus, using the \texttt{multi-qa-mpnet} teacher model results in a performance decline. 
Importantly, the Greek part of this dataset has been translated by \textsc{PhilTa}$_{\text{En} \rightarrow \text{Grc}}$, which could possibly have affected the quality of the dataset. Indeed, while we see this negative trend when \emph{testing} on the generated data, the inclusion of the \philta-generated Ancient Greek Rosenthal corpus as additional \emph{training data} leads to a notable enhancement for the Greek datasets, while the performance for Latin translation retrieval remains largely unaffected.

The results for Latin-to-Greek and Greek-to-Latin translations are shown in \cref{tab:resultsgrcla}. Our models notably exhibit strong performance across both datasets. Again, utilizing the Greek Rosenthal data considerably improves performance.
These results show that \sphilberta can be efficiently utilized in a scenario that solely involves Greek and Latin texts, without necessitating the involvement of English texts.

\label{sec:appendix}
\begin{table*}
    \centering
    \resizebox{\linewidth}{!}{
    \begin{tabular}{p{10.85cm}p{.5cm}p{15cm}}
         \textbf{Query} && \textbf{Results} \\
         \midrule
         \textit{\colorbox{BurntOrange}{Haec} ubi \colorbox{Dandelion}{dicta} dedit, \colorbox{SkyBlue}{lacrimantem} et multa volentem} \newline \colorbox{BurntOrange}{This} \colorbox{Dandelion}{speech} uttered, while I \colorbox{SkyBlue}{wept} and would have said many a thing, && \textgreek{τῆς δ᾽ ἄρ᾽ \colorbox{Dandelion}{ἀκουούσης} \colorbox{SkyBlue}{ῥέε δάκρυα}, τήκετο δὲ χρώς·} \newline and as she \colorbox{Dandelion}{listened} her \colorbox{SkyBlue}{tears flowed} and her face melted \newline \textgreek{\colorbox{BurntOrange}{ὣς} \colorbox{Dandelion}{φάτο}, τῆς δ᾽ εὔνησε \colorbox{SkyBlue}{γόον}, σχέθε δ᾽ ὄσσε \colorbox{SkyBlue}{γόοιο}.} \newline \colorbox{BurntOrange}{So} she \colorbox{Dandelion}{spoke}, and lulled Penelope's \colorbox{SkyBlue}{laments}, and made her eyes to cease from \colorbox{SkyBlue}{weeping}. \newline \textgreek{\colorbox{BurntOrange}{ὣς} \colorbox{Dandelion}{φάτο}, τῇ δ᾽ ἄρα \colorbox{SkyBlue}{θυμὸν} ἐνὶ στήθεσσιν \colorbox{SkyBlue}{ὄρινε}.} \newline  \colorbox{BurntOrange}{So} he \colorbox{Dandelion}{spoke}, and \colorbox{SkyBlue}{stirred the heart} in her breast.\\
         &&\\
         \textit{\colorbox{BurntOrange}{dicere deseruit}, \colorbox{Dandelion}{tenuisque recessit in auras}.} \newline \colorbox{BurntOrange}{[...said]}, she left me and \colorbox{Dandelion}{retreated into thin air}.&& \textgreek{ἡ μὲν ἄρ᾽ ὣς ἔρξασ᾽ \colorbox{Dandelion}{ἀπεβήσετο δῖα θεάων},} \newline Now when she had done this \colorbox{Dandelion}{the fair goddess departed}, \newline \textgreek{ἡ μὲν ἄρ᾽ \colorbox{BurntOrange}{ὣς εἰποῦσ᾽} \colorbox{Dandelion}{ἀπέβη πρὸς δώματα καλά},} \newline \colorbox{BurntOrange}{So saying}, \colorbox{Dandelion}{she departed to the fair palace}. \newline \textgreek{ἡ μὲν ἄρ᾽ \colorbox{Dandelion}{ἐς κρήνην κατεβήσετο καλλιρέεθρον}} \newline [She] had \colorbox{Dandelion}{come down to the fair-flowing spring} [Artacia], \\
        &&\\
        \textit{\colorbox{BurntOrange}{Ter} \colorbox{Dandelion}{conatus} ibi collo dare \colorbox{SkyBlue}{bracchia} circum:} \newline \colorbox{BurntOrange}{Thrice} there was I fain to lay mine \colorbox{SkyBlue}{arms} round her neck; && \textgreek{ὄπτ᾽ ἐν \colorbox{SkyBlue}{χερσὶν} ἑλών, τά ῥά οἱ γέρα πάρθεσαν αὐτῷ.} \newline he took in his \colorbox{SkyBlue}{hands} roast meat and set it before them, [...] which they had set before himself as a mess of honor. \newline \textgreek{\colorbox{BurntOrange}{τρὶς} μέν μιν πελέμιξεν ἐρύσσεσθαι \colorbox{Dandelion}{μενεαίνων},} \newline \colorbox{BurntOrange}{Thrice} he made it quiver \colorbox{Dandelion}{in his eagerness} to draw it, \newline \textgreek{αὐτίκ᾽ ἔπειτα \colorbox{BurntOrange}{τρίαιναν} ἑλὼν \colorbox{SkyBlue}{χερσὶ} στιβαρῇσιν} \newline straightway took his \colorbox{BurntOrange}{trident} in his mighty \colorbox{SkyBlue}{hands}, \\
        &&\\
        \textit{\colorbox{BurntOrange}{ter} frustra \colorbox{SkyBlue}{comprensa} \colorbox{JungleGreen}{manus effugit} \colorbox{Dandelion}{imago},} \newline \colorbox{BurntOrange}{thrice} \colorbox{Dandelion}{the vision} I vainly \colorbox{SkyBlue}{clasped}  \colorbox{JungleGreen}{fled out of my hands},&& \textgreek{\colorbox{BurntOrange}{τρὶς} δέ μοι \colorbox{JungleGreen}{ἐκ χειρῶν} \colorbox{Dandelion}{σκιῇ εἴκελον ἢ καὶ ὀνείρῳ}} \newline 
        and \colorbox{BurntOrange}{thrice} [she flitted] \colorbox{JungleGreen}{from my arms} like \colorbox{Dandelion}{a shadow or a dream}, \newline
        \textgreek{\colorbox{BurntOrange}{τρὶς} μὲν ἐφωρμήθην, \colorbox{SkyBlue}{ἑλέειν} τέ με θυμὸς ἀνώγει,}  \newline \colorbox{BurntOrange}{Thrice} I sprang towards her, and my heart bade me \colorbox{SkyBlue}{clasp} her, \newline \textgreek{\colorbox{JungleGreen}{χερσὶ} δὲ μή τι λίην προκαλίζεο, μή με χολώσῃς,} \newline  But with thy \colorbox{JungleGreen}{[hands]} do not provoke me overmuch, \\
        &&\\
        \textit{\colorbox{BurntOrange}{par levibus ventis volucrique} simillima \colorbox{Dandelion}{somno}.} \newline \colorbox{BurntOrange}{even as the light breezes}, or most like to fluttering \colorbox{Dandelion}{sleep}. && \textgreek{ἡ δ᾽ ἔθεεν \colorbox{BurntOrange}{Βορέῃ ἀνέμῳ} \colorbox{BurntOrange}{ἀκραέϊ καλῷ,}} \newline And she ran before the \colorbox{BurntOrange}{North Wind}, \colorbox{BurntOrange}{blowing fresh and fair}, \newline \textgreek{ὄρσας ἀργαλέων \colorbox{BurntOrange}{ἀνέμων} ἀμέγαρτον \colorbox{BurntOrange}{ἀυτμήν},} \newline  when he had roused a furious \colorbox{BurntOrange}{blast} of cruel \colorbox{BurntOrange}{winds} \newline \textgreek{ἐς \colorbox{BurntOrange}{πνοιὰς ἀνέμων}. ἡ δ᾽ ἐξ \colorbox{Dandelion}{ὕπνου} ἀνόρουσε} \newline into the \colorbox{BurntOrange}{breath of the winds}. And [she] started up from \colorbox{Dandelion}{sleep}\\
    \end{tabular}}
    \caption{Top 3 predictions of our best-performing \sphilberta model (teacher: \texttt{multi-qa-mpnet}; student: \philberta; additional \philta-generated Rosenthal data) when queried over the whole \textit{Odyssey}. We mark corresponding cross-lingual concept pairs with individual colors.}
    \label{tab:examples}
\end{table*}

\section{Case Study: The \textit{Aeneid} and Homer's \textit{Odyssey}}
Examinations of the intertextual allusions in Virgil's \textit{Aeneid} to both the \textit{Iliad} and the \textit{Odyssey} have a long history, dating back to antiquity.  Structurally, the \textit{Aeneid's} initial six books mirror the narrative of the \textit{Odyssey}, while the concluding six books correspond more closely to the \textit{Iliad}. 

In the second book of the \textit{Aeneid}, the protagonist Aeneas attempts to escape from the ravaged city of Troy with his family. Tragically, his wife, Creusa, is lost amidst the chaos. Creusa's ghost consoles him and bids him goodbye before receding into thin air: \enquote{\textit{This speech uttered, while I wept and would have said many a thing, she left me and retreated into thin air. Thrice there was I fain to lay mine arms round her neck; thrice the vision I vainly clasped fled out of my hands, even as the light breezes, or most like to fluttering sleep.}}\footnote{Virgil, \textit{Aeneid}, 2.790--794, translated by \citet{mackail1885aeneid}.} 

These verses mirror closely a scene in the Nekyia of the \textit{Odyssey}, where Odysseus meets his mother Anticleia in the underworld: \enquote{\textit{So she spoke, and I pondered in heart, and was fain to clasp the spirit of my dead mother. Thrice I sprang towards her, and my heart bade me clasp her, and thrice she flitted from my arms like a shadow or a dream, and pain grew ever sharper at my heart.}}\footnote{Homer, \textit{Odyssey}, 11.204--208, translated by \citet{odyssey}.} 

To evaluate our model's proficiency in identifying these intertextual allusions, we employ each verse of the \textit{Aeneid} passage (i.e., 5 verses) as a query, which we then compare to the verse embeddings (approx. \numprint{11000} verses) of the complete \textit{Odyssey}. \cref{tab:examples} shows
the three highest results for each verse,
according to our best-performing model setup (teacher: \texttt{multi-qa-mpnet}; student: \philberta; additional \philta-generated Rosenthal data).

We note that these verses do not share a direct one-to-one relationship and they are not translations of each other, the scenario in which our model was trained. Even so, we observe that verse 793 (\enquote{\textit{thrice the vision I vainly clasped fled out of my hands}}) is correctly paired with the best corresponding Greek verse (\enquote{\textit{and thrice she flitted from my arms like a shadow or a dream}}). In the majority of cases, our model accurately captures crucial concepts, such as \textit{weeping}, \textit{departing}, \textit{triplicity}, \textit{wind}, and \textit{sleep}, linking them reasonably to different verses. However, our verse-to-verse mapping, which precludes longer texts, results in a lack of a cohesive concept of extended intertextually alluding passages.

Still, our case study demonstrates the proficiency of our models 
in recognizing sentence structures and translating them to a different language (as in \enquote{\textit{this speech uttered}} $\rightarrow$ \enquote{\textit{so she spoke}}), and in identifying common topics or concepts across languages, even locating verses where multiple relevant concepts exist within the same verse (\enquote{\textit{thrice}}, \enquote{\textit{the vision}}, \enquote{\textit{out of my hands}}  $\rightarrow$ \enquote{\textit{thrice}}, \enquote{\textit{a shadow or a dream}}, \enquote{\textit{from my arms}}).

Despite these remarkable results, our case study 
also reveals the need for a more sophisticated retrieval mechanism that extends beyond verse boundaries to search for broader patterns. 
Yet, already in the present state, our \sphilberta model can serve as a useful tool for automatic first-pass exploration of potential cross-lingual intertextual allusions, and in this way can support philologists in the search for intertextual allusions. 

\section{Conclusion}
We introduce \sphilberta, a multilingual \philberta-derived sentence transformer model, specifically adapted to Classical Philology. 
Our model represents a pioneering effort in detecting intertextual allusions between Ancient Greek and Latin texts, which is characteristic of many Roman writers who used Greek literature for juxtaposition. \sphilberta displays impressive performance across various datasets, confidently identifying direct translations among English, Latin, and Ancient Greek. 
We have illustrated that \sphilberta holds strong potential in revealing intertextual allusions; however, additional research is needed to fully exploit the model's capabilities. Our multilingual \sphilberta and the similarity-driven retrieval settings built upon it offer, for the first time, the option to study intertextuality cross-lingually on a broader scale in original Classical Literature. 

\section*{Acknowledgements}
We thank Nina Stahl for the insightful discussions we shared on the Greek and Latin parallel passages.

\FloatBarrier
\bibliography{anthology,custom}
\bibliographystyle{acl_natbib}

\appendix
\end{document}